\documentclass[sigconf,10pt,nonacm,screen]{acmart}
\setcopyright{none}
\usepackage{amsfonts}
\usepackage{bbm}
\usepackage{booktabs}
\usepackage{makecell}
\usepackage{tikz}
\usetikzlibrary{arrows.meta}
\usepackage{pgfplots}
\pgfplotsset{compat=1.18}
\usetikzlibrary{shapes.geometric, patterns, positioning}
\usepackage{adjustbox}
\usepackage{enumitem}
\usepackage{wrapfig}
\usepackage{subcaption}
\usepackage{xspace}
\usepackage[svgnames]{xcolor}

\usepackage{enumitem}

\usepackage[switch*]{lineno}
\makeatletter
\def\@LN@column{2}
\makeatother

\makeatletter
\if@ACM@review
  \def\ACM@mk@linecount{%
    \savebox{\ACM@linecount@bx}[4em][t]{\parbox[t]{4em}{\normalfont
        \normalsize
        \setlength{\ACM@linecount@bxht}{0pt}%
        \loop{\color{LightGray}\scriptsize\the\ACM@linecount}\\
        \global\advance\ACM@linecount by \@ne
        \addtolength{\ACM@linecount@bxht}{\baselineskip}%
        \ifdim\ACM@linecount@bxht<\textheight\repeat
        {\color{LightGray}\scriptsize\the\ACM@linecount}\hfill
        \global\advance\ACM@linecount by \@ne}}}
\fi
\makeatother

\usepackage{algorithm}
\usepackage[noend]{algpseudocode}
\usepackage{mathrsfs}
\usepackage{amsmath}

\usepackage{amssymb}
\usepackage[makeroom]{cancel}
\usepackage{amsthm}

\usepackage{booktabs}

\usepackage{colortbl}

\usepackage[subtle]{savetrees}
\usepackage{array}      
\usepackage{booktabs}   
\usepackage{breqn}
\usepackage{tabularx}

\usepackage[framemethod=tikz]{mdframed}
\newmdenv[
    linecolor=gray,
    linewidth=1pt,
    topline=false,
    bottomline=false,
    rightline=false,
    skipabove=2pt,
    skipbelow=1pt,
    leftmargin=0,
    rightmargin=10pt,
    innerleftmargin=3pt,
    innerrightmargin=0pt,
    innertopmargin=1pt,
    innerbottommargin=2pt,
    backgroundcolor=white
]{txtframe}
\newmdenv[
    linecolor=gray,
    linewidth=1pt,
    roundcorner=4pt,
    skipabove=2pt,
    skipbelow=1pt,
    leftmargin=0,
    rightmargin=0,
    innerleftmargin=6pt,
    innerrightmargin=6pt,
    innertopmargin=5pt,
    innerbottommargin=5pt,
    backgroundcolor=gray!10
]{graytxtframe}
\newmdenv[
    linecolor=gray!40,
    linewidth=0.5pt,
    roundcorner=3pt,
    skipabove=2pt,
    skipbelow=2pt,
    leftmargin=0pt,
    rightmargin=0pt,
    innerleftmargin=6pt,
    innerrightmargin=6pt,
    innertopmargin=3pt,
    innerbottommargin=3pt,
    backgroundcolor=gray!12
]{insightframe}

\usepackage[utf8]{inputenc}
\usepackage[T1]{fontenc}

\newif\ifshowcomment
\showcommenttrue

\newcommand{\sys}{{\textsc{TypoNet}}\xspace}

\newcommand{\tocite}[1]{{\textcolor{red}{\textbf{[~]}}}}
\newcommand{\toref}[1]{\textcolor{red}{\textbf{N}}}

\newcommand{\eg}{\emph{e.g.,} }
\DeclareRobustCommand{\txtsl}[1]{%
  \ifmmode
    \text{\fontsize{9}{10}\selectfont\fontencoding{T1}\fontfamily{lmr}\fontshape{sl}\selectfont #1}%
  \else
    {\fontsize{9}{10}\selectfont\fontencoding{T1}\fontfamily{lmr}\fontshape{sl}\selectfont #1}%
  \fi}
\makeatletter
\newcommand{\etc}{etc\@ifnextchar.{}{.\@\xspace}}
\makeatother

\ifshowcomment
\newcommand{\TODO}[1]{\textcolor{red}{{[\small\textsf{{TODO: #1}}}]}}
\newcommand{\NOTE}[1]{\textcolor{orange}{{[\small\textsf{{NOTE: #1}}}]}}
\else
\newcommand{\TODO}[1]{}
\newcommand{\NOTE}[1]{}
\fi

\newcommand{\remove}[1]{}

\newcommand{\mypar}[1]{{\noindent\bf #1.\ }}

\newcommand{\circleblack}[1]{%
 \begin{tikzpicture}[baseline=(char.base)]
   \node[draw,circle,inner sep=0.5pt, fill=black, text=white] (char){\small #1};
 \end{tikzpicture}%
 }

 \usepackage{tcolorbox}
\tcbuselibrary{listings}
\usepackage{graphicx}

\newtcolorbox{promptbox}{
  colback=gray!5,
  colframe=gray!60,
  listing only,
  listing options={
    basicstyle=\ttfamily\small,
    breaklines=true,
    columns=fullflexible
  }
}

\AtBeginDocument{%
  \setlength{\textfloatsep}{6pt plus 1pt minus 1pt}%
  \setlength{\floatsep}{6pt plus 1pt minus 1pt}%
  \setlength{\intextsep}{6pt plus 1pt minus 1pt}%
  \setlength{\dbltextfloatsep}{6pt plus 1pt minus 1pt}%
  \setlength{\abovecaptionskip}{3pt}%
  \setlength{\belowcaptionskip}{0pt}%
  \setlength{\abovedisplayskip}{4pt plus 1pt minus 1pt}%
  \setlength{\belowdisplayskip}{4pt plus 1pt minus 1pt}%
  \setlength{\abovedisplayshortskip}{2pt plus 1pt}%
  \setlength{\belowdisplayshortskip}{2pt plus 1pt}%
}

\title[Let AI Agents Translate Networks]{\fontsize{19pt}{19pt}\selectfont Let AI Agents Translate Networks, Not Reason About Them}
\subtitle{Typographical Network Modeling with Automated and Verifiable Axiomatization}

\author{Hongyu H\`e}
\affiliation{%
  \institution{Princeton University}%
  \country{}%
}

\author{Maria Apostolaki}
\affiliation{%
  \institution{Princeton University}%
  \country{}%
}

\makeatletter
\AtBeginMaketitle{%
  \apptocmd{\@mkauthors}{%
    \global\setbox\mktitle@bx=\vbox{%
      \unvbox\mktitle@bx
      \centering
      \normalsize\@date\par
      \medskip
    }%
  }{}{}%
}
\makeatother

\begin{document}

\begin{abstract}
A formal model enables verifying reachability, localizing an outage, or anticipating the blast radius of a change.
Yet, virtually no production network has one, since writing a model by hand demands rare expertise and is hard to keep current as the network changes frequently.
At its core, network modeling is a typographical exercise: it translates network artifacts (\eg configurations, topology, and routing state) into rules in formal logic.
Translation of this kind is what large language models (LLMs) nowadays do well.
Unlike free-form AI reasoning, such translation can be formally verified.

Once modeling is no longer the bottleneck, trusting AI to reason over large, complex networks no longer makes sense.
Our position therefore cuts against the prevailing race to put autonomous AI agents in charge end-to-end.
We instead confine AI to translation and rely on a solver for reliable long-horizon reasoning, building a reusable formal model of general network behavior that can then be specialized to specific tasks, \eg root-cause analysis (RCA).
We build \sys that constructs and validates a symbolic model of an emulated production-scale WAN from the network's own artifacts.
Our preliminary evaluation shows \sys helps in two ways.
On its own, \sys answers operational questions (\eg reachability verification and change-impact analysis) faster, more cheaply, and more reliably than an LLM.
As a tool for an AI agent, \sys boosts fault localization at lower cost.
The result makes the case for AI that builds verifiable network models and relies on a solver for reliable long-horizon reasoning.
\end{abstract}


\maketitle

\section{Introduction} \label{sec:intro}

\begin{figure}[t]
\centering
\begin{adjustbox}{width=\linewidth,center=0pt}
\includegraphics[width=\linewidth]{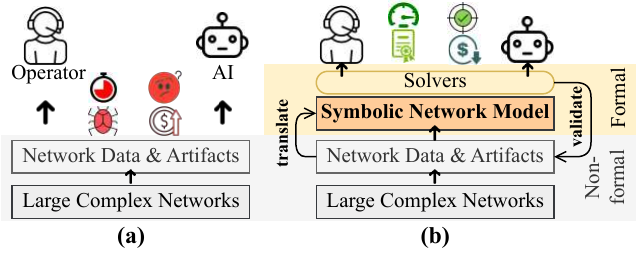}
\end{adjustbox}
\caption{How large networks are operated today, and what reliable operation actually needs.
\textbf{(a) Status quo.} Working directly from raw artifacts is slow and error-prone for operators and treacherous for AI agents: the data volume overflows the context window, cascading hallucinations derail reasoning, and reprocessing the same data each time is costly.
\textbf{(b) \sys.} It \emph{translates} the artifacts into rules in formal logic and \emph{validates} each against the network's own evidence, yielding a symbolic model that operators and AI agents query by offloading formal reasoning to a solver.}
\label{fig:intro}
\end{figure}

Running a large network is an exercise in answering questions about how it behaves.
An operator must know whether a service is still reachable, which links a customer's traffic crosses, what a planned configuration change will break, and, when an outage strikes, which device caused it.
Answering them precisely requires a formal model of the network, a machine-checkable description of how its configurations, topology, and routing state combine to forward packets~\cite{batfish2015, minesweeper2017, hsa2012, veriflow2012}.
Given such a formal model, a solver can verify reachability, localize a fault, or predict the blast radius of a change; without one, operators fall back on ad hoc scripts and tribal knowledge.
Yet for almost every production network no such model exists, because building one by hand requires rare expertise in both formal methods and networking, demands prohibitive manual efforts, and is hard to keep current as the network changes frequently~\cite{config2spec2020, modelfreeverification2025}.

The current trend in the community is to skip the formal model and put an AI agent in charge.
The underlying hope is AI agents that watch the live network, reason about what is wrong, and act to repair it end to end~\cite{hamadanian2023llmincident, mani2023enhancing}.
Large language models (LLMs) sit at the center of that ambition, and a fast-growing line of work asks them to configure, diagnose, and manage networks directly~\cite{wu2024_netllm, nada2024_designing, han2024_netconfeval}.
For critical network infrastructure, this approach is misplaced, because it trusts the LLM where it is weakest.
LLMs hallucinate, and their errors compound over long chains of reasoning, so a single confident misstep derails an entire diagnosis~\cite{farquhar2024hallucinate, zhang2023hallucinate, huang2025hallucinate, dziri2023faith}.
They also cannot hold a hyperscale network in view, since millions of devices and hundreds of regions do not fit in any context window.
Accuracy degrades further as the relevant facts recede into a long prompt~\cite{liu2024lost, levy2024same}.
Fault localization, the heart of troubleshooting, is exactly where cascading errors and missing context do the most damage.
Fig.~\ref{fig:intro}a depicts the regime we distrust, in which an AI agent works directly from the raw artifacts of a large network and its accuracy degrades as the network grows.

This paper makes a different bet: do not ask the LLM to reason about the network at all.
Use it to build a formal model that a solver can check, and let the solver do the complex, long-horizon reasoning.
The LLM is confined to one job: translating the network's heterogeneous artifacts into logical rules.
\sys then tests each rule against independent network evidence before trusting it.
Two guarantees then stand apart.
The solver is sound with respect to the symbolic model, so it correctly derives what the rules entail.
When a rule is wrong, the refutation that exposes it names the specific axiom at fault, so an error localizes to a single rule the loop can then repair.
The symbolic model's faithfulness to the real network is a separate matter, since it holds only as far as the evidence that tried and failed to refute it.
Core reasoning moves from the LLM to the solver, and the symbolic model becomes the subject under scrutiny.
\emph{The LLM translates; the solver decides.}

The bet is principled, because a network is a manmade, typographical artifact.
Its configurations, topology records, the logical rules describing its behavior, and the program a solver runs are all strings.
Building a formal model is therefore a translation among those symbol systems.
\emph{A translation can be checked in a way that open-ended AI reasoning cannot.}
LLMs are the strongest translators between symbol systems we have~\cite{hotnets23_routerconfigs, gong2023fmml}.
The one risky step, writing the formal model, thus becomes one we can validate.

We realize the idea in \sys, which constructs and validates a symbolic model of a network from its existing artifacts.
The construction loop is adversarial, inspired by counterexample-guided inductive synthesis (CEGIS)~\cite{solar-lezama2006sketch}. 
An LLM-driven Constructor proposes logical rules, each a formal statement about one facet of how the network behaves.
A Detractor refutes them with counterexamples drawn from the network's own evidence, \eg a reachability measurement that contradicts a proposed rule.
The loop repeats until the symbolic model withstands attack and satisfies the network's established invariants, the properties a correct network must uphold in every valid state.
Each rule is \emph{axiomatized} bottom-up, built on simpler facts and states the loop has already validated.
Behavior therefore grows from ground facts into the multi-hop relations operators ask about.
We call such a validated set of rules a \emph{theory}: a conjunction of rules, closed under entailment~\cite{he2026netnomos,sathandbook2021}.
A solver reasons over the theory to answer questions about the network's behavior.
\sys builds the symbolic model \emph{compositionally}, in layers: a reusable foundation theory of general network behavior, then vendor- and deployment-specific refinements, and finally per-task specializations.
Root-cause analysis (RCA) is one such specialization: we teach the foundation theory how faults manifest by injecting them and modeling the resulting cause-symptom links.

We prototype \sys on an emulated WAN of tens of autonomous systems (ASes), scaled to approximate a production deployment's thousands of core devices.
The symbolic model checks the invariants operators rely on, \eg all-pairs reachability and policy compliance.
Used as a tool by an AI agent, the symbolic model localizes injected faults with a verified solver query at near-zero token cost~\cite{josephson1996abductive, calcagno2009biabduction}.
We measure the resulting gains in token cost and fault-localization accuracy against LLM-only baselines and SOTA agentic RCA solutions.
The same symbolic model can also serve reachability verification, change-impact analysis, and configuration-drift detection, so it is built once and reused across tasks~\cite{config2spec2020, relnetverif2024}.
AI agents keep a role as users of formal models, and the reasoning we have to trust runs on the solver.
Fig.~\ref{fig:intro}b shows the paradigm we advocate, in which operators and AI agents alike offload formal reasoning to a solver that runs over automatically constructed symbolic models.

We argue that automatically constructing a formal model turns network modeling from a bespoke, expert-driven craft into an automated and verifiable process.
We give preliminary evidence on three fronts.
The construction loop converges into a foundation theory that matches held-out network behavior and transfers almost unchanged to a WAN deployment twice as large.
Once built, the theory answers operational questions such as blast radius in milliseconds and at no token cost.
As a tool for RCA, it sharpens fault localization for every frontier AI agent we test, and lifts a small, inexpensive model to competitive accuracy at a fraction of a frontier agent's cost.
We close with the open questions that decide how far the idea reaches, from keeping the symbolic model synchronized under constant network change to certifying the network models an AI creates.
\section{Background and Motivation} \label{sec:motivation}

\begin{figure*}[t]
\centering
\begin{adjustbox}{width=\linewidth,center=0pt}
\includegraphics[width=\linewidth]{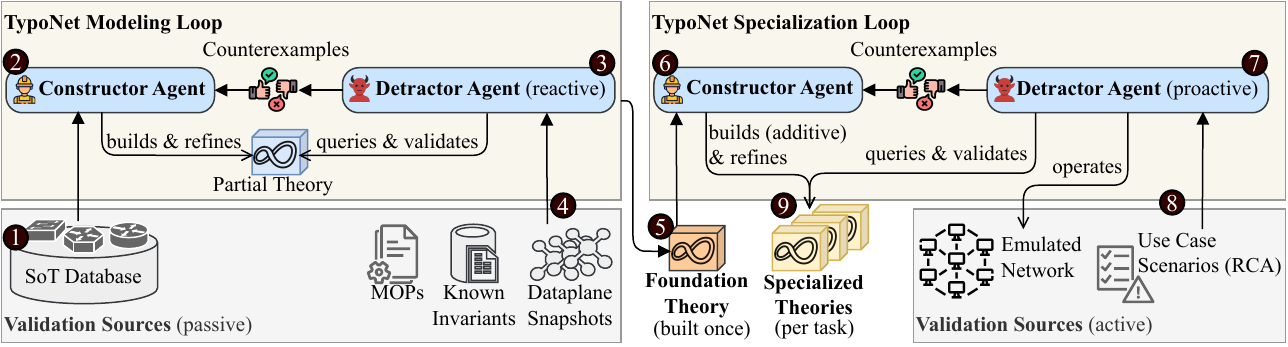}
\end{adjustbox}
\caption{\sys's two agentic CEGIS loops turn a network's own records into a verifiable symbolic model.
A \emph{Modeling Loop} (left) builds a reusable Foundation Theory from passive evidence, and a \emph{Specialization Loop} (right) adds thin per-task layers validated on an emulated network.}
\label{fig:overview}
\end{figure*}

Modern network operations already run on a written record of the network.
Every large operator maintains a source-of-truth (SoT) database that stores, as typed and linked records, each device, interface, link, address block, and routing session, together with the roles and policy intent behind them.
Meta's Robotron~\cite{robotron2016} and Google's MALT~\cite{mogul2020malt} are two examples, and comparable systems run at other hyperscalers~\cite{lyu2024poseidon}.
Provisioning, monitoring, and configuration generation all read from that record, which makes it the closest thing a network has to a single authoritative description.
Three trends now make that record the natural starting point for an automatically built formal model.

\mypar{SoT databases capture structure, but leave semantics implicit}
A SoT database records what the network is made of, but leaves how it behaves unstated.
For instance, it states that two routers share a routing session and that an interface carries an access list.
It does not explicitly say whether a destination is reachable, which path a flow takes, which flows a given link carries, or what breaks when a device fails.
Properties of this kind are entailed by the records yet never written in a form a machine can compute over, \eg reachability, waypointing, impact cones, blast radius, and the causal links between a fault and its symptoms.
Blast radius is a representative example: before every deployment operators must estimate the reach of a change.
A single edit to one core device can cascade many hops away, and a wrong estimate risks an outage~\cite{crescent2024, gill2011understanding, mahajan2002bgp}.
Predicting that reach reliably needs long-horizon reasoning operators can trust, which is where a sound formal model, and not a manual or LLM guess, earns its place.
The barrier is semantic: a SoT stores ground facts but lacks the rules, entailment, and recursion that behavior requires.
\emph{The behavior is latent in the records but never expressed in a computable form.}
Surfacing it as a checkable formal model is the gap \sys bridges.

\mypar{Formal methods stall on the hand-built model}
Two decades of research can verify reachability, synthesize configurations, and localize faults once a formal model exists~\cite{batfish2015, minesweeper2017, netkat2014, propane2016, arzani2018007}.
The hard part is obtaining that model.
Existing tools assume a hand-crafted translation from raw artifacts into formal semantics, which demands scarce dual expertise and decays as the network changes~\cite{config2spec2020}.
Hand-crafting also cannot keep up with scale.
A single hyperscale network spans tens of millions of devices and hundreds of datacenters~\cite{validatingdatacenters2019, crystalnet2017, govindan2016evolve}, and its records change continuously as configurations, failures, and repairs land.
A hand-built model is therefore stale the moment it is finished, so operators report that verification tooling sees little real-world use~\cite{modelfreeverification2025}.
Only an automated method can build a formal model at that scale and keep it synchronized~\cite{crescent2024, s2verifier2025, krentsel2024validateinputs, crosscheck2026}.

\mypar{LLMs have become good enough at translation}
The missing capability was a way to produce the formal model automatically, cheaply, and without having to trust its author.
LLMs now convert among configurations, intents, and formal artifacts with enough fidelity to be useful.
Because a translation can be checked and corrected, their mistakes need not be believed~\cite{mani2023enhancing, hotnets23_routerconfigs, gong2023fmml}.
Mature solvers stand ready to consume millions of variables and reason over them soundly~\cite{z3, sathandbook2021, sat_scaling}.
For the first time, the translator and the reasoner co-exist.

\mypar{\sys differs from prior tools in where the semantics come from}
Two other lines of work also turn a network into something a machine can reason about, and \sys parts from both in the origin of its semantics.
Configuration-parsing verifiers such as Batfish derive behavior by parsing vendor configurations through hand-written parsers that encode each vendor's semantics~\cite{batfish2015, minesweeper2017}.
Those vendor parsers carry the whole translation, stay hand-coded, and are never checked against the network's own evidence.
An error in a parser therefore becomes a silent error in every answer.
Tool-using AI agents go the other way and let the LLM reason over the live network directly, calling tools to fetch state~\cite{hamadanian2023llmincident, mani2023enhancing}.
The semantics then live inside the LLM's reasoning, no step is validated, and the least reliable component is trusted the most.
\sys instead keeps the semantics in an explicit symbolic model, synthesizes it automatically, and validates every rule against independent network evidence before a solver reasons over it.


\section{Automated \& Verifiable Network Modeling} \label{sec:approach}

\sys constructs a formal model of a network automatically, and it makes the symbolic model trustworthy even though an LLM writes it.
Fig.~\ref{fig:overview} shows how the two agentic loops reconcile these goals end to end, and five principles, stated below, make the design work.

\mypar{Build the symbolic model from the network's own records}
A production network already documents itself in a SoT database (\S\ref{sec:motivation}), so the raw material for a symbolic model exists before \sys runs.
\sys treats each such record as a ground fact~\circleblack{1} and asks the LLM to do one thing only: compose those facts into the rules that explain how the network behaves~\circleblack{2}.
Bottom-up \emph{axiomatization} gives the symbolic model three layers: ground facts at the bottom~\circleblack{1}, rules that derive operational states from them, and rules that combine those states into multi-hop relations.
Because each rule is defined only over facts and states already validated beneath it, definitions never turn circular, and \sys validates the symbolic model one layer at a time.
Grounding the symbolic model in existing records removes the manual labor, since no expert transcribes the network and the LLM only translates.

\mypar{Refute a rule before trusting it}
A rule is only as trustworthy as our ability to attack it.
\sys pairs the LLM Constructor with an adversarial Detractor (\circleblack{3}) that refutes each proposed rule against what the network actually does.
A rule the evidence refutes is discarded however plausible it reads, so the network's own evidence is the validation oracle that keeps the symbolic model tied to the real network.
Every refutation returns a counterexample the Constructor must repair (\S\ref{sec:intro}), and the loop stops once a full validation pass raises no new counterexample.
The Detractor draws on two kinds of evidence: \emph{passive} sources (\circleblack{4}) already recorded about the network, and \emph{active} sources (\circleblack{8}) it produces on demand from an emulated network.
\sys's trust base is therefore small and explicit: the network's own evidence, an emulated network, and the solver.
The LLM's output is never part of it, since a rule it proposes counts for nothing until the evidence corroborates it.

\mypar{Ground the foundation in operators' invariants and procedures}
The reactive Detractor builds the Foundation Theory from passive sources alone.
Reproducing every recorded case is not enough, since a symbolic model can match them all and still break on the one that matters.
\sys therefore also grades the symbolic model against the network's known invariants (\S\ref{sec:intro}), the laws every deployed network upholds.
Such invariants include all-pairs reachability, loop freedom, and header-precise access-control compliance~\cite{nod2015, validatingdatacenters2019}.
A second source is the Method of Procedures (MOPs) operators run for planned changes, whose steps are virtually all validated in production~\cite{fastbgpsim, symmetrysurgery2016, crystalnet2017}.
Operators author the invariants and MOPs independently of the symbolic model, so satisfying them is external evidence that the symbolic model reflects the real network.

\mypar{Compose the symbolic model from a reusable foundation and thin refinement layers}
A network model should cover many vendors, deployment conditions, and downstream tasks, more than any single hand-built artifact can carry.
\sys therefore takes a \emph{compositional} approach.
It builds a Foundation Theory (\circleblack{5}) of general network behavior once, then \emph{composes} thin refinement layers onto it (\circleblack{6}) for a given deployment's policy and a given vendor's quirks.
The symbolic model is thus an assembly of independently built and independently validated theories (\circleblack{9}).
The composition works because of how the rules are written: each rule quantifies over facts and names no specific devices.
The Foundation Theory therefore states behavior every IP network shares, \eg how forwarding follows the installed routes, without reference to any one topology.
Specializing to a concrete network is then mostly a matter of supplying that network's facts, since \sys loads the new SoT under the unchanged rules.
Only a genuinely new behavior, such as a vendor feature the foundation does not cover, needs a fresh layer.
The same reusable core then serves many downstream applications, \eg reachability verification, change-impact analysis, and root-cause analysis.

\mypar{Confine specialization to an emulated network}
The proactive Detractor~(\circleblack{7}) builds the specialization layers, and here it does more than react.
It actively drives an emulated network through use-case scenarios (\circleblack{8}) and reads the resulting states, because the knowledge it needs does not exist until a scenario is run.
Learning how a network behaves under a fault means causing that fault.
\sys therefore confines every perturbing operation to an \emph{emulated network} and lets production contribute read-only artifacts alone.
For RCA, \sys injects a fault, observes which invariants break, and reverts it.
It records the resulting fault-to-symptom relationship as a new specialization layer, which lets the symbolic model reason backward from an incident's symptoms to the smallest set of faults that explain them.
The emulated network need not be a full copy of production: a high-fidelity digital twin or a smaller network that exercises the same behaviors suffices.
Such emulation at this scale is common practice in production~\cite{miniinternet2020, kollaps2020, s2verifier2025, mirrornet, modelfreeverification2025}.

An emulated network alone does not replace the symbolic model: it runs only forward, so it shows what a fault does but cannot invert an observed symptom into its cause the way RCA requires~\cite{crystalnet2017, crescent2024, sat_scaling}.
The emulated network is thus the offline apparatus \sys learns from. 
The symbolic model, on the other hand, is the lightweight, synchronized artifact operators query online.

Operators query the finished symbolic model by entailment.
A query $q$ holds under \emph{cautious} entailment, that is, when $q$ is true in every stable model of the theory \texttt{Th}.
Concretely, $\texttt{Th} \models q$, which \sys decides on the theory's stratified fragment by refutation, checking that $\texttt{Th} \land \neg q$ has no stable model.
Each answer is a sound chain of rule applications: it starts from ground facts, derives operational states, and composes them into the multi-hop relations operators ask about.
One query can therefore traverse many devices and hops no operator could follow by hand.
\section{Preliminary Results} \label{sec:eval}


\mypar{Implementation}
We implement \sys in Answer Set Programming (ASP), executed by a proprietary backend in the spirit of Network-Optimized Datalog~\cite{nod2015} for performance.
ASP's support for default negation and abductive inference suits not only property checking but also the backward, cause-seeking reasoning RCA needs~\cite{josephson1996abductive, aliseda2006abductive, walton2014abductive}.

\mypar{Setup}
We deliberately build the foundation theory and the specialized theory on different networks, so the experiment tests both the feasibility and the transferability of \sys's compositional modeling.
We build the foundation theory on a 30-AS emulated WAN of 515 devices.
We then transfer that theory to a more than $2\times$ larger 70-AS deployment of 1{,}191 devices and specialize it there, a scale on par with the core of a production WAN~\cite{crosscheck2026, krentsel2024validateinputs, crescent2024, s2verifier2025}.

\subsection{Constructing Valid Foundation Theory} \label{sec:eval:construction}

The construction loop converges to a Foundation Theory at a one-time, offline cost, and it agrees with held-out behavior (Table~\ref{tab:construction}).
The Detractor is what ties the theory to evidence: disable it, and the Constructor still produces plausible-looking rules, but the theory admits vacuous rules that fire on no real state.
As a negative control, we plant deliberately wrong rules by mutating a correct rule's predicate or direction, and the Detractor catches nearly all of them.

\begin{table}[t]
\centering
\small
\setlength{\tabcolsep}{5pt}
\renewcommand{\arraystretch}{1.1}
\begin{tabular}{@{}lr@{}}
\toprule
Theory construction measurement & Value \\
\midrule
Foundation rules (LLM-authored / human-seeded) & 128 (121 / 7) \\
Counterexamples raised by the Detractor & 214 \\
Refutation rounds per rule (avg.) & 2.8 \\
One-time construction cost & 3\,h, \$18.6 \\
Held-out agreement with ground truth & $97.4\%$ \\
\quad without the Detractor (ablation) & $41\%$ \\
Planted-wrong rules caught (negative control) & 48 / 50 \\
Rules reused unchanged, 30-AS $\rightarrow$ 70-AS & 118 / 128 ($92\%$) \\
\bottomrule
\end{tabular}
\caption{Adversarial refutation is what makes the automatically built Foundation Theory trustworthy. Disabling the Detractor collapses held-out agreement 58\%; once validated, $92\%$ of the rules transfer unchanged to a $2\times$ larger deployment.}
\label{tab:construction}
\end{table}

\subsection{Answering Operational Questions} \label{sec:eval:queries}

Once validated, the foundation theory answers operational questions with no LLM in the loop.
It chains many facts across devices and hops, where a manual check or a free-form LLM loses track of a cascading consequence.
A single query decides all-pairs reachability, for instance, and pinpoints the pair and hop at which it breaks.
Blast radius is a representative question, because one change at a single device propagates through a long chain of consequences.
As a real example, an operator plans to take a core interface out of service, modeled as $\neg\,\txtsl{IsUp}(\texttt{spineB}, \texttt{p17})$, where switch \texttt{swX} prefers the uplink \texttt{eth2} toward it and holds \texttt{eth1} as a backup for the video class, while the control class has none.
\sys applies the change and unrolls the cascade one validated axiom at a time:
\begin{align*}
&\neg\,\txtsl{IsUp}(\texttt{spineB}, \texttt{p17}) \land \txtsl{ConnectsTo}(\texttt{swX}, \texttt{eth2}, \texttt{spineB}, \texttt{p17}) \\
&\quad \models \neg\,\txtsl{IsUp}(\texttt{swX}, \texttt{eth2}) 
\ \models \neg\,\txtsl{CanForward}(\texttt{swX}, \texttt{eth2}, \texttt{c\_vid}), \\
&\txtsl{Preferred}(\texttt{swX}, \texttt{c\_vid}, \texttt{eth2})
  \land \txtsl{Backup}(\texttt{swX}, \texttt{c\_vid}, \texttt{eth1}) \\
&\quad \land\, \txtsl{CanForward}(\texttt{swX}, \texttt{eth1}, \texttt{c\_vid}) \\
&\quad \models \txtsl{IsRedirectedTo}(\texttt{swX}, \texttt{c\_vid}, \texttt{eth1}), \\
&\txtsl{IsRedirectedTo}(\texttt{swX}, \texttt{c\_vid}, \texttt{eth1})
  \land \txtsl{DropRate}(\texttt{swX}, \texttt{eth1}, \texttt{high}) \\
&\quad \models \txtsl{IsDegraded}(\texttt{p\_vid}), \\
&\neg\,\exists i\!:\ \txtsl{CanForward}(\texttt{swX}, i, \texttt{c\_ctrl}) \\
&\quad \models \neg\,\txtsl{IsReachable}(\texttt{swX}, \texttt{p\_ctrl})
\ \models \neg\,\txtsl{IsReachable}(\texttt{torA}, \texttt{p\_ctrl}).
\end{align*}
The single change fans out into a \emph{full impact cone}: the video prefix \texttt{p\_vid} survives but degrades on the overloaded backup, while the control prefix \texttt{p\_ctrl}, which had no backup, goes dark and drags an upstream top-of-rack down with it.
Estimating that reach by hand or by an LLM's guess is the error-prone, long-horizon step that motivates \sys~\cite{crescent2024}.
In practice, the same chained reasoning also drives alerting, where \sys elevates a log only when it entails a degraded prefix or lost reachability.

\mypar{Per-query cost}
Each query runs on the solver alone with no LLM in the loop, so its cost is a few milliseconds and no tokens at all.
\sys answers four representative queries over the running example (healthy and post-change reachability, blast radius, and the video class's waypoint), each correct and within $5$--$34$\,ms.
The blast-radius query alone chains 23 rules across 11 hops, whereas a single-shot LLM given the same ground facts clears only the one-hop reachability case.

\subsection{Agentic RCA with a Specialized Theory} \label{sec:eval:rca}

We specialize the foundation theory into a fault-to-symptom layer and give it to an AI agent as a tool for root-cause analysis.
To evaluate it, we port the NIKA open benchmark~\cite{nika2025} into our emulated WAN and adopt its scoring exactly as the SADE agent does~\cite{sade2026}.
Each AI agent inspects a live incident and names the faulty devices and the fault type.

\mypar{Method}
We split incidents into train and test by fault type and topology region, so no fault family or region seen during specialization reappears at test time.
The test set holds 96 held-out incidents and 40 healthy controls that a trustworthy AI agent should leave alone.
Every configuration runs under one shared prompt and tool schema, repeated 5 times, and we report the average values.

\begin{figure}[t]
\centering
\definecolor{cgpt}{HTML}{2563EB}
\definecolor{copus}{HTML}{DC2626}
\definecolor{csonnet}{HTML}{EAB308}
\definecolor{csonnetdk}{HTML}{A16207}
\definecolor{csade}{HTML}{16A34A}
\tikzset{
  mrk/.style={draw, line width=0.7pt, inner sep=0pt},
  gptbl/.style={mrk, circle, minimum size=6.5pt, draw=cgpt, fill=white},
  gpttn/.style={mrk, circle, minimum size=6.5pt, draw=cgpt, fill=cgpt},
  gptsd/.style={mrk, circle, minimum size=6.5pt, draw=cgpt, pattern=north east lines, pattern color=cgpt},
  opbl/.style={mrk, rectangle, minimum size=6.5pt, draw=copus, fill=white},
  optn/.style={mrk, rectangle, minimum size=6.5pt, draw=copus, fill=copus},
  opsd/.style={mrk, rectangle, minimum size=6.5pt, draw=copus, pattern=north east lines, pattern color=copus},
  sotn/.style={mrk, regular polygon, regular polygon sides=3, minimum size=8.5pt, draw=csonnetdk, fill=csonnet},
  sobl/.style={mrk, regular polygon, regular polygon sides=3, minimum size=8.5pt, draw=csonnetdk, fill=white},
  sadea/.style={mrk, diamond, minimum size=8pt, draw=csade, pattern=north east lines, pattern color=csade},
  lgtypo/.style={mrk, rectangle, minimum size=6.5pt, draw=gray!35!black, fill=gray!45!black},
  lgbase/.style={mrk, rectangle, minimum size=6.5pt, draw=gray!35!black, fill=white},
  lgsade/.style={mrk, rectangle, minimum size=6.5pt, draw=gray!35!black, pattern=north east lines, pattern color=gray!35!black},
  lt/.style={anchor=west, font=\scriptsize, inner sep=1pt},
  lh/.style={anchor=west, font=\scriptsize\bfseries, inner sep=1pt},
}
\resizebox{\columnwidth}{!}{%
\begin{tikzpicture}
\begin{axis}[
  name=ax,
  width=8.6cm, height=6.0cm,
  x dir=reverse,
  xlabel={Cost per Diagnosis [USD] (cheaper $\rightarrow$)},
  ylabel={Localization F1 ($\uparrow$)},
  xmin=0.45, xmax=1.14, ymin=0.30, ymax=0.80,
  xtick={0.5,0.6,0.7,0.8,0.9,1.0,1.1}, ytick={0.3,0.4,0.5,0.6,0.7,0.8},
  grid=both, major grid style={dotted, gray!55},
  tick label style={font=\small}, label style={font=\small},
  axis line style={gray!70}, clip=false,
]
\node[gptbl] at (axis cs:0.563,0.672){};
\node[gpttn] at (axis cs:0.675,0.757){};
\node[gptsd] at (axis cs:0.754,0.506){};
\node[opbl] at (axis cs:0.979,0.529){};
\node[optn] at (axis cs:0.970,0.578){};
\node[opsd] at (axis cs:1.069,0.529){};
\node[sotn] at (axis cs:0.50,0.511){};
\node[sobl] at (axis cs:0.527,0.41){};
\node[sadea] at (axis cs:0.637,0.369){};
\draw[-{Stealth[length=1.8mm]}, cgpt, line width=0.6pt, shorten <=3.5pt, shorten >=3.5pt, opacity=0.75] (axis cs:0.563,0.672) -- (axis cs:0.675,0.757);
\draw[-{Stealth[length=1.8mm]}, copus, line width=0.6pt, shorten <=3.5pt, shorten >=3.5pt, opacity=0.75] (axis cs:0.979,0.529) -- (axis cs:0.970,0.578);
\draw[-{Stealth[length=1.8mm]}, csonnetdk, line width=0.6pt, shorten <=4pt, shorten >=4pt, opacity=0.85] (axis cs:0.527,0.41) -- (axis cs:0.50,0.511);
\draw[-{Triangle[length=4mm,width=5.5mm]}, line width=4pt, black!30, line cap=round] (axis cs:1.10,0.345) -- (axis cs:1.005,0.45);
\node[anchor=north, font=\small\bfseries, black!30] at (axis cs:1.05,0.359){Better};
\end{axis}
\node[lh, anchor=west] (mlab) at ([yshift=10mm]ax.north west) {Model:};
\node[gpttn, right=4pt of mlab] (s1){};   \node[lt, right=2pt of s1] (t1){GPT-5.6 Sol};
\node[optn, right=7pt of t1] (s2){};      \node[lt, right=2pt of s2] (t2){Opus 4.8};
\node[sotn, right=7pt of t2] (s3){};      \node[lt, right=2pt of s3] (t3){Sonnet 4.6};
\node[sadea, right=7pt of t3] (s4){};     \node[lt, right=2pt of s4] (t4){SADE agent};
\node[lh, anchor=west] (slab) at ([yshift=4mm]ax.north west) {Skills:};
\node[lgtypo, right=4pt of slab] (u1){};  \node[lt, right=2pt of u1] (v1){+TypoNet};
\node[lgbase, right=7pt of v1] (u2){};    \node[lt, right=2pt of u2] (v2){Baseline};
\node[lgsade, right=7pt of v2] (u3){};    \node[lt, right=2pt of u3] (v3){+SADE skills};
\end{tikzpicture}%
}
\caption{\sys's specialized theory improves fault localization for every LLM. GPT-5.6 Sol with \sys is the most accurate at cost below any Opus 4.8 configuration, and Sonnet 4.6 with \sys is the cheapest point above $0.5$ localization F1 on the accuracy/cost frontier.}
\label{fig:rca_tradeoff}
\end{figure}
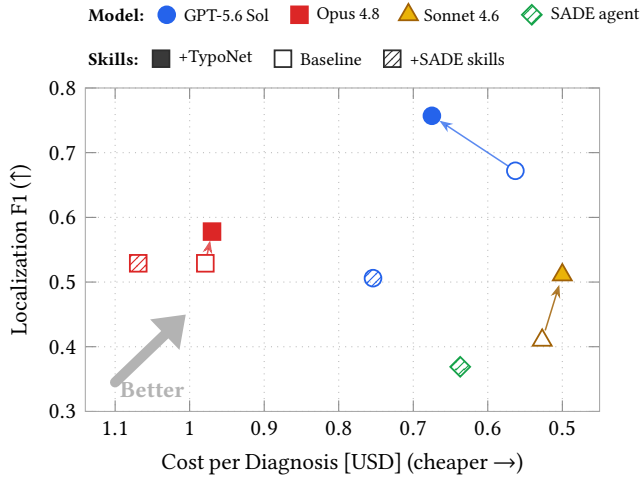

\mypar{Configurations}
We run three frontier LLMs, GPT-5.6 Sol, Opus 4.8, and the smaller Sonnet 4.6, each unaided (baseline) and each equipped with the \sys tool.
We add the SADE skill library on the two frontier LLMs and the SADE agent itself, for nine configurations in all.
Every configuration faces the same held-out incidents and the same healthy controls, and Fig.~\ref{fig:rca_tradeoff} plots each one's localization F1 against its cost per run.

\begin{insightframe}
\noindent\textbf{Insight 1:} Even frontier AI agents localize faults more accurately and raise fewer false alarms with \sys's formal reasoning.
\end{insightframe}
Adding the \sys tool to the same LLM helps both frontier AI agents: GPT-5.6 Sol's localization F1 climbs from $0.672$ to $0.757$, while Opus 4.8 gains most in naming.
The edge widens on the hardest incidents, where the symptom surfaces far from its cause: there \sys lifts localization F1 from $0.59$ to $0.82$ and roughly doubles Opus 4.8's.
A matched ablation on GPT-5.6 Sol isolates the source of the gain, holding the prompt, tool schema, and call budget fixed.
The theory rendered as plain text reaches F1~$0.710$, a randomly shuffled theory falls to $0.55$, and the validated theory behind a solver reaches $0.757$.
The gain therefore comes from the validated formal layer and the solver that uses it.
The same formal grounding also curbs false alarms: the baseline frontier AI agents flag a fault on almost \emph{every} healthy network scenario.
Using \sys as a tool cuts that false-alarm rate by $2$--$3\times$.
This observation serves as a preliminary warning for anyone using an AI agent for monitoring.

\begin{insightframe}
\noindent\textbf{Insight 2:} With \sys, a small, less powerful LLM reaches a competitive accuracy/cost point.
\end{insightframe}
On its own Sonnet 4.6 localizes poorly (F1~$0.41$), but adding \sys raises it to $0.511$ at just \$0.50 per run.
It undercuts the GPT-5.6 Sol baseline and roughly halves the per-run cost of the Opus 4.8 baseline at comparable localization.
The uplift is largest on the hardest incidents, where Sonnet 4.6 with \sys reaches F1~$0.83$ and clears both \emph{unaided} frontier baselines.
Naming stays the small LLM's weak point.
\section{Research Agenda} \label{sec:agenda}


\mypar{Keep the symbolic model in step with a network that changes frequently}
A production WAN is reconfigured continuously, so a symbolic model correct when built drifts within hours.
A stale symbolic model is worse than none, since it can certify a property that the live network no longer satisfies.
Synchronization is therefore a first-class requirement~\cite{robotron2016, crescent2024}.
Since most changes touch a small part of the network, the open question is recompiling only the affected layers soundly and fast enough to track the network in near real time.

\mypar{Turn vendor-specific behavior into theory automatically}
Vendor-specific features cause over $30\%$ of production failures~\cite{crescent2024, s2verifier2025}, yet a general Foundation Theory leaves them out.
The proactive Detractor can learn them by driving an emulated network through scenarios that exercise a feature.
Each scenario surfaces gaps between the symbolic model's prediction and the emulator's behavior, which the Detractor turns into new rules.
Two questions stay open: which scenarios expose a vendor's key behaviors, and how a learned layer transfers across deployments on the same platform.

\mypar{Confront the incompleteness of established knowledge}
The established knowledge our Foundation Theory draws on is abundant but underspecified, written in prose that is often silent on corner cases.
The theory is therefore inevitably incomplete, and specialization narrows the gap without ever closing it.
A deeper obstacle is that observation does not imply causation: a mined fault-to-symptom relationship is only a correlation between an injected condition and its symptom.
Isolating the true cause needs controlled experiments that vary one factor at a time, itself hard at network scale.
Until we bound what the theory does not know, a verified answer holds only relative to the current theory.

\mypar{Reach up the stack and fold in telemetry}
\sys today mostly models Layer 2 and Layer 3 forwarding, leaving higher layers outside the theory.
Stateful middleboxes such as NAT, firewalls, and load balancers are the first stress test, since a stateless forwarding theory cannot express their connection state.
Datalog-based verification already reaches host-level reachability~\cite{nod2015}, so the same logical style can climb the stack.
A second frontier is numeric telemetry: congestion and SLA violation would extend a representation like MALT~\cite{mogul2020malt} with numeric facts and rules.
Both raise fidelity while enlarging the solver's search, so which properties earn their cost remains an open question.


\newpage
\bibliographystyle{ACM-Reference-Format}
\bibliography{reference}

\end{document}